# Solving Hybrid Influence Diagrams with Deterministic Variables

**Yijing Li and Prakash P. Shenoy**

University of Kansas, School of Business
1300 Sunnyside Ave., Summerfield Hall
Lawrence, KS 66045-7601, USA
{*yjl, pshenoy*}@ku.edu

## Abstract

We describe a framework and an algorithm for solving hybrid influence diagrams with discrete, continuous, and deterministic chance variables, and discrete and continuous decision variables. A continuous chance variable in an influence diagram is said to be deterministic if its conditional distributions have zero variances. The solution algorithm is an extension of Shenoy's fusion algorithm for discrete influence diagrams. We describe an extended Shenoy-Shafer architecture for propagation of discrete, continuous, and utility potentials in hybrid influence diagrams that include deterministic chance variables. The algorithm and framework are illustrated by solving two small examples.

## 1 Introduction

An influence diagram (ID) is a formal compact representation of a Bayesian decision-making problem. It consists of four parts: a sequence of decisions, a set of chance variables with a joint distribution represented by a hybrid Bayesian network (BN), the decision maker's preferences for the uncertain outcomes represented by a joint utility function that factors additively, and information constraints that indicate which uncertainties are known (and which are unknown) when a decision has to be made. IDs were initially defined by [8].

Hybrid IDs are IDs containing a mix of discrete and continuous chance and decision variables. In practice, most decision problems are hybrid. However, solving a hybrid ID involves two main computational challenges. First, marginalizing a continuous chance variable involves integration of a product of density and utility functions. In some cases, such as the Gaussian density function, there may not exist a closed-form representation of the integral. We will refer to this problem as the *integration* problem.

Second, marginalizing a decision variable involves maximizing a utility function. If the decision variable is continuous and has relevant continuous information predecessors, then we may be faced with the problem of finding a closed-form solution of the maximization problem. Not only do we have to find the maximum value of the decision variable (as a function of the states of its relevant information predecessors), we have also to find a closed-form expression of the maximum utility (as a function of the states of its relevant information predecessors). We will refer to this problem as the *optimization* problem.

A traditional method for solving hybrid IDs is to approximate a hybrid ID with a discrete ID by discretizing the continuous chance and decision variables (see, e.g., [9]). If we discretize a continuous variable using a few bins, we may have an unacceptable approximation of the problem. On the other hand, if we use many bins, we increase the computational effort of solving the resulting discrete ID. In the BN literature, [10] describes a dynamic non-uniform discretization technique for chance variables depending on where the posterior density lies. This technique needs to be adapted for solving hybrid IDs.

Another method for solving hybrid IDs is to use Monte Carlo (MC) simulation. One of the earliest to suggest MC methods for solving decision trees was [6], where the entire joint distribution is sampled. [2] proposes a MC method that samples from a small set of chance variables at a time for each decision variable. [13] proposes several MC methods and provide bounds on the number of samples required given some error bounds. [1] explores using Markov chain MC methods to solve a single-stage decision problem with continuous decision and chance nodes to solve the maximization problem.

Among exact methods, [16] provides a theory to solve

IDs where all chance and decision variables are continuous. The continuous chance variables have conditional linear Gaussian (CLG) distributions, and the utility function is quadratic. Such IDs are called Gaussian IDs. These assumptions ensure that the joint distribution of all chance variables is multivariate normal whose marginals can be easily found without the need for integration. Also, the quadratic nature of the utility function ensures that there is a unique maximum than can be computed in closed form without the need for solving an optimization problem.

[14] extends Gaussian IDs to include discrete chance variables that do not have continuous parents. If a continuous chance variable does not have a CLG distribution, then it can be approximated by a mixture of Gaussians represented by a discrete variable with mixture weights and a continuous variable with the discrete variable as its parent and with a CLG distribution.

To find marginals in hybrid BNs, [12] proposes approximating probability density functions (PDF) by mixtures of truncated exponentials (MTE) as a solution for the integration problem. The family of MTE functions is easy to integrate, is closed under combination and marginalization, and can be propagated using the Shenoy-Shafer architecture [18]. [4] describes MTE IDs, where the PDFs of continuous chance variables, and utility functions, are described using MTE functions. Thus any PDF can be used as long as they can be approximated by MTE. Discrete variables can have continuous parents, and there is no restriction on the nature of the utility function.

Similar to MTE, [21] proposes approximating PDFs by piecewise polynomial functions called mixture of polynomials (MOP). Like MTE, MOP functions are closed under multiplication and integration. Thus, they can be used to find exact marginals in hybrid BNs. MOP functions have some advantages over MTE functions. MOP approximations can be easily found using the Taylor series for differentiable functions, even for multidimensional functions. Also, they are closed for a larger class of deterministic functions than MTE functions, which are closed only for linear deterministic functions.

[3] describes arc reversals in hybrid BNs that contain a mix of discrete, continuous, and deterministic variables. The conditionals for discrete variables are represented by discrete potentials, for continuous variables by density potentials, and for deterministic variables by Dirac potentials (containing a weighted sum of Dirac delta functions [5]). The deterministic function does not have to be linear or invertible. The only requirement is that it should be differentiable. [20] extends this framework by defining mixed potentials, and combination marginalization operations for mixed potentials. They call their framework the extended Shenoy-Shafer architecture.

In this paper, we propose further extending the extended Shenoy-Shafer architecture for solving hybrid IDs that include deterministic chance variables. The solution technique is conceptually the same as the one proposed by [17] for discrete decision problems. We extend this method to include continuous and deterministic chance variables, and continuous decision variables. To address the integration problem, we propose MOP approximations of PDFs and utility functions. Since MOP functions are easily integrable, finding the maximum of a utility function that is in MOP form is also easier.

An outline of the remainder of this paper is as follows. In section 2, we describe the framework and an algorithm to solve hybrid IDs with deterministic variables. In section 3, we describe MOP functions. In section 4, we solve two small decision problems to illustrate our framework and algorithm. Finally, in section 5, we conclude with a summary and discussion on the limitations of our algorithm.

## 2 The Extended Shenoy-Shafer Framework

In this section, we describe a framework and an algorithm for solving hybrid IDs with deterministic variables. The algorithm described is adapted from [17, 11]. The framework described here is a further extension of the extended Shenoy-Shafer architecture described in [20] for inference in hybrid Bayesian networks with deterministic variables. Here, we include decision variables, and utility potentials. Mixed potentials are a triple of discrete, continuous, and utility potentials. The definition of marginalization of mixed potentials is revised to accommodate non-associativity of multiplication and addition in the definition of combination of potentials.

**Variables and States.** We are concerned with a finite set $\mathcal{V} = \mathcal{D} \cup \mathcal{C}$ of *variables*. Variables in $\mathcal{D}$ are called *decision* variables, and variables in $\mathcal{C}$ are called *chance* variables. Each variable $X \in \mathcal{V}$ is associated with a set $\Omega_X$ of possible *states*. If $\Omega_X$ is finite or countable, we say $X$ is *discrete*, otherwise $X$ is *continuous*. We will assume that the state space of continuous variables is the set of real numbers, and that the state space of discrete variables is a set of symbols (not necessarily real numbers). If $r \subseteq \mathcal{V}, r \neq \emptyset$, then $\Omega_r = \times \{\Omega_X | X \in r\}$. If $r = \emptyset$, we will adopt the convention that $\Omega_\emptyset = \{\diamond\}$.

We will distinguish between discrete chance variables and continuous chance variables. Let $\mathcal{C}_d$ denote the set of all discrete chance variables and let $\mathcal{C}_c$ denote the set of all continuous chance variables. Then, $\mathcal{C} = \mathcal{C}_d \cup \mathcal{C}_c$. We do not distinguish between discrete and continuous decision variables.

In an ID, each chance variable has a conditional distribution function for each state of its parents. A conditional distribution function associated with a continuous variable is said to be *deterministic* if the variances (for each state of its parents) are all zeros. For simplicity, henceforth, we will refer to continuous variables with non-deterministic conditionals as continuous, and continuous variables with deterministic conditionals as *deterministic*.

In an ID, we will depict decision variables by rectangular nodes, discrete chance variables by single-bordered elliptical nodes, continuous chance variables by double-bordered elliptical nodes, deterministic chance variables by triple-bordered elliptical chance nodes, and additive factors of the joint utility function by diamond-shaped nodes. We do not distinguish between discrete and continuous decision variables.

**Projection of States.** If $\mathbf{x} \in \Omega_r$, $\mathbf{y} \in \Omega_s$, and $r \cap s = \emptyset$, then $(\mathbf{x}, \mathbf{y}) \in \Omega_{r \cup s}$. Thus, $(\mathbf{x}, \diamond) = \mathbf{x}$. Suppose $\mathbf{x} \in \Omega_r$, and $s \subseteq r$. Then, the *projection* of $\mathbf{x}$ to $s$, denoted by $\mathbf{x}^{\downarrow s}$, is the state of $s$ obtained from $\mathbf{x}$ by dropping states of $r \setminus s$. Thus, e.g., $(w, x, y, z)^{\downarrow \{W, X\}} = (w, x)$, where $w \in \Omega_W$, and $x \in \Omega_X$. If $s = r$, then $\mathbf{x}^{\downarrow s} = \mathbf{x}$. If $s = \emptyset$, then $\mathbf{x}^{\downarrow s} = \diamond$.

**Discrete Potentials.** In an ID, the conditional probability functions associated with chance variables are represented by functions called *potentials*. If $A$ is discrete, it is associated with a conditional probability mass function. The conditional probability mass functions are represented by functions called *discrete potentials*. Suppose $r \subseteq \mathcal{V}$ is such that if it is non-empty then it contains a discrete variable. A discrete potential for $r$ is a function $\alpha : \Omega_r \to [0, 1]$. The values of discrete potentials are probabilities.

Although the domain of the potential $\alpha$ is $\Omega_r$, we will refer to $r$ as the *domain* of $\alpha$. Thus, the domain of a potential representing the conditional probability function associated with some chance variable $X$ in an ID is always the set $\{X\} \cup pa(X)$, where $pa(X)$ denotes the set of parents of $X$ in the ID graph. Notice that a discrete potential can have continuous chance variables or decision variables in its domain, but, in this case, it must also include a discrete chance variable, i.e., if the domain of a discrete potential is non-empty, then it must include a discrete variable. The values of discrete potentials are always in units of probability.

**Density Potentials.** Continuous chance variables are typically associated with conditional PDFs. Conditional PDFs are represented by functions called *density potentials*. Suppose $r \subseteq \mathcal{V}$ is such that if it is non-empty then it contains a continuous variable. A density potential $\zeta$ for $r$ is a function $\zeta : \Omega_r \to \mathbb{R}^+$, where $\mathbb{R}^+$ is the set of non-negative real numbers. The values of density potentials are probability densities. Notice that a density potential can have discrete chance variables or decision variables in its domain, but, in this case, it must include a continuous chance variable, and its values are always in units of (probability) density.

**Dirac Delta Functions.** Deterministic variables have conditional distributions described by equations. We will represent such distributions by *Dirac potentials* that use Dirac delta functions $\delta$ [5].

$\delta : \mathbb{R} \to \mathbb{R}^+$ is called a Dirac delta function if $\delta(x) = 0$ if $x \neq 0$, and $\int \delta(x) dx = 1$. Whenever the limits of integration of an integral are not specified, the entire range $(-\infty, \infty)$ is to be understood. $\delta$ is not a proper function since the value of the function at 0 doesn't exist (i.e., is not finite). It can be regarded as a limit of a certain sequence of functions (such as, e.g., the Gaussian density function with mean 0 and variance $\sigma^2$ in the limit as $\sigma \to 0$). However, it can be used as if it were a proper function for practically all our purposes without getting incorrect results. It was first defined by Dirac [5].

**Dirac Potentials.** Suppose $t = r \cup s$ is a set of variables containing some discrete variables $r$ and some continuous chance variables $s$, and suppose $s \neq \emptyset$. A *Dirac potential* for $t$ is a function $\xi : \Omega_t \to \mathbb{R}^+$ such that $\xi(\mathbf{r}, \mathbf{s})$ is of the form:

$$\xi(\mathbf{r}, \mathbf{s}) = \Sigma \{ p_{\mathbf{r},i} \delta(z - g_{\mathbf{r},i}(\mathbf{s}^{\downarrow (s \setminus \{Z\})})) | \mathbf{r} \in \Omega_r, \\ i = 1, \ldots, n_{\mathbf{r}} \}, \quad (1)$$

where $\mathbf{r} \in \Omega_r$, $\mathbf{s} \in \Omega_s$, $Z \in s$ is a continuous or deterministic variable, $z \in \Omega_Z$, $\delta(z - g_{\mathbf{r},i}(\mathbf{s}^{\downarrow (s \setminus \{Z\})}))$ are Dirac delta functions, $p_{\mathbf{r},i}$ are probabilities for all $i = 1, \ldots, n_{\mathbf{r}}$, and $n_{\mathbf{r}}$ is a positive integer. Here, we are assuming that continuous or deterministic variable $Z$ is a weighted sum of deterministic functions $g_{\mathbf{r},i}(\mathbf{s}^{\downarrow (s \setminus \{Z\})})$ of the other continuous variables in $s$, with probability weights $p_{\mathbf{r},i}$, and that the nature of the deterministic functions and weights may depend on the state $\mathbf{r}$ of the discrete variables in $r$, or on some latent index $i$. Like density potentials, Dirac potentials must include a continuous variable in its domain if it is non-empty, and its values are in units of density.

**Continuous Potentials.** Both density and Dirac potentials are special instances of a broader class of potentials called continuous potentials. Suppose $t \subseteq \mathcal{V}$ is such that if it is non-empty then it contains a continuous chance variable. Then, a *continuous potential* for $t$ is a function $\xi : \Omega_t \to \mathbb{R}^+$. Like density and Dirac potentials, continuous potentials must include a continuous variable in its domain if it is non-empty, and its values are in units of density.

**Utility Potentials.** An ID representation includes utility functions, that represent the preferences of the decision maker for the various outcomes. If an ID has more than one utility node, we assume an additive factorization of the joint utility function. Each additive factor of the utility function is represented by a utility potential. A utility potential $\upsilon$ for $t \subseteq \mathcal{V}$ is a function $\upsilon : \Omega_t \to \mathbb{R}$. The values of utility potentials are in units of utiles.

**Mixed Potentials.** To keep track of the nature of potentials, we define mixed potentials. A mixed potential has three parts. The first part is a discrete potential, the second is a continuous potential, and the third is a utility potential. Suppose $\alpha$ is a discrete potential for $r$. Then, a mixed potential representation of $\alpha$ is $\mu = (\alpha, \iota_c, \iota_u)$, where $\iota_c$ denotes the identity continuous potential for $\emptyset$, $\iota_c(\diamond) = 1$, and $\iota_u$ denotes the identity utility potential for $\emptyset$, $\iota_u(\diamond) = 0$. Suppose $\zeta$ is a continuous potential for $s$. Then, a mixed potential representation of $\zeta$ is $\mu = (\iota_d, \zeta, \iota_u)$, where $\iota_d$ denotes the discrete identity potential for $\emptyset$, $\iota_d(\diamond) = 1$. Finally, if $\upsilon$ is a utility potential for $t$, then a mixed potential representation of $\upsilon$ is $\mu = (\iota_d, \iota_c, \upsilon)$. $\iota_d$ is a discrete potential for $\emptyset$ whose value is in units of probability, $\iota_c$ is a continuous potential for $\emptyset$ whose value is in units of density, and $\iota_u$ is an utility potential for $\emptyset$ whose value is in units of utiles.

**Combination of Potentials.** The definition of combination of potentials depends on the nature (discrete, continuous, or utility) of potentials. Although there are nine permutations, we have only two distinct definitions. Utility functions are additive factors of the joint utility function. Thus, combination of two utility potentials involves pointwise addition. In all other eight cases, combination of potentials involves pointwise multiplication.

Suppose $\upsilon_1$ is a utility potential for $t_1$ and $\upsilon_2$ is a utility potential for $t_2$. Then, the combination of $\upsilon_1$ and $\upsilon_2$, denoted by $\upsilon_1 \otimes \upsilon_2$, is a utility potential for $t_1 \cup t_2$ given by:

$$(\upsilon_1 \otimes \upsilon_2)(\mathbf{x}) = \upsilon_1(\mathbf{x}^{\downarrow t_1}) + \upsilon_2(\mathbf{x}^{\downarrow t_2}) \quad \text{for all } \mathbf{x} \in \Omega_{t_1 \cup t_2}. \tag{2}$$

Suppose $\alpha_1$ is a potential (discrete, continuous, or utility) for $t_1$ and $\alpha_2$ is a potential (discrete, continuous, or utility) for $t_2$. Suppose that both $\alpha_1$ and $\alpha_2$ are not both utility. Then, the combination of $\alpha_1$ and $\alpha_2$, denoted by $\alpha_1 \otimes \alpha_2$, is a potential for $t_1 \cup t_2$ given by:

$$(\alpha_1 \otimes \alpha_2)(\mathbf{x}) = \alpha_1(\mathbf{x}^{\downarrow t_1}) \alpha_2(\mathbf{x}^{\downarrow t_2}) \quad \text{for all } \mathbf{x} \in \Omega_{t_1 \cup t_2}. \tag{3}$$

If $\alpha_1$ and $\alpha_2$ are both discrete, then $\alpha_1 \otimes \alpha_2$ is a discrete potential. If $\alpha_1$ and $\alpha_2$ are both continuous, then $\alpha_1 \otimes \alpha_2$ is a continuous potential. If $\alpha_1$ is discrete or continuous, and $\alpha_2$ is utility, or vice-versa, then $\alpha_1 \otimes \alpha_2$ is a utility potential. In all other cases, we will define the nature of the the combined potential when we define marginalization of mixed potentials.

**Combination of Mixed Potentials.** Suppose $\mu_1 = (\alpha_1, \zeta_1, \upsilon_1)$ and $\mu_2 = (\alpha_2, \zeta_2, \upsilon_2)$ are two mixed potentials for $r_1 \cup s_1 \cup t_1$ and $r_2 \cup s_2 \cup t_2$, respectively, with discrete parts $\alpha_1$ for $r_1$, and $\alpha_2$ for $r_2$, respectively, continuous parts $\zeta_1$ for $s_1$, and $\zeta_2$ for $s_2$, respectively, and utility parts $\upsilon_1$ for $t_1$, and $\upsilon_2$ for $t_2$, respectively. Then, the combination of $\mu_1$ and $\mu_2$, denoted by $\mu_1 \otimes \mu_2$, is a mixed potential for $(r_1 \cup s_1 \cup t_1) \cup (r_2 \cup s_2 \cup t_2)$ given by

$$\mu_1 \otimes \mu_2 = (\alpha_1 \otimes \alpha_2, \zeta_1 \otimes \zeta_2, \upsilon_1 \otimes \upsilon_2) \tag{4}$$

Since the combination of two discrete potentials is discrete, two continuous potentials is continuous, and two utility potentials is utility, the definition in eq. (4) is consistent with the definition of mixed potentials. If $\mu_1 = (\alpha, \iota_c, \iota_u)$ is a mixed potential for $r$, $\mu_2 = (\iota_d, \zeta, \iota_u)$ is a mixed potential for $s$, and $\mu_3 = (\iota_d, \iota_c, \upsilon)$ is a mixed potential for $t$, then $\mu_1 \otimes \mu_2 \otimes \mu_3 = (\alpha, \zeta, \upsilon)$ is a mixed potential for $r \cup s \cup t$.

It is easy to confirm that combination of mixed potentials is commutative and associative. This follows from the commutativity and associativity of the combination of discrete potentials, the combination of continuous potentials, and the combination of utility potentials.

**Marginalization of Potentials.** The definition of marginalization of potentials depends on the nature of the variable being marginalized. We marginalize discrete chance variables by addition over its state space, continuous chance variables by integration over its state space, and decision variables (discrete or continuous) by maximization over its state space.

Suppose $\alpha$ is a potential (discrete or continuous or utility or some combination of these) for $a$, and suppose $X \in a$ is a discrete variable. Then, the marginal of $\alpha$ by deleting $X$, denoted by $\alpha^{-X}$, is a potential for $a \setminus \{X\}$ given as follows:

$$\alpha^{-X}(\mathbf{y}) = \Sigma \{\alpha(x, \mathbf{y}) | x \in \Omega_X\} \quad \text{for all } \mathbf{y} \in \Omega_{a \setminus \{X\}}. \tag{5}$$

If $X \in a$ is a continuous variable, then $\alpha^{-X}$ is defined as follows:

$$\alpha^{-X}(\mathbf{y}) = \int_{-\infty}^{\infty} \alpha(x, \mathbf{y})dx \quad \text{for all } \mathbf{y} \in \Omega_{a \setminus \{X\}}. \quad (6)$$

And if $X \in a$ is a decision variable, then $\alpha^{-X}$ is defined as follows:

$$\alpha^{-X}(\mathbf{y}) = \max\{\alpha(x, \mathbf{y}) | x \in \Omega_X\} \quad \text{for all } \mathbf{y} \in \Omega_{a \setminus \{X\}}. \quad (7)$$

In some examples, if $X$ is a decision variable, the state space of $X$ may be further constrained as a function of the states of some of its information predecessors. In this case, we assume that the maximization in eq. (7) is subject to these additional constraints.

**Division of Potentials.** The definition of marginalization of mixed potentials involves division of probability (discrete or continuous) potentials by probability potentials. Also, the potential in the divisor is always a marginal of the potential being divided.

Suppose $\alpha$ is a discrete or continuous potential for $a$, and suppose $X \in a$ is a discrete or continuous chance variable. Then the division of $\alpha$ by $\alpha^{-X}$, denoted by $\alpha \oslash \alpha^{-X}$, is a potential for $a$ defined as follows:

$$(\alpha \oslash \alpha^{-X})(x, \mathbf{y}) = \alpha(x, \mathbf{y}) / \alpha^{-X}(\mathbf{y}) \quad (8)$$

for all $x \in \Omega_X$, and $\mathbf{y} \in \Omega_{a \setminus \{X\}}$. In eq. (8), if the denominator is zero, then the numerator is also 0, and in this case we define $0/0$ as 0. The nature of the potential $\alpha \oslash \alpha^{-X}$ depends on $X$. If $X$ is discrete, $\alpha \oslash \alpha^{-X}$ is discrete, and if $X$ is continuous, $\alpha \oslash \alpha^{-X}$ is continuous [3].

**Marginalization of Mixed Potentials.** Similar to marginalization of potentials, marginalization of mixed potentials depends on the nature of the variable being marginalized. We distinguish between marginalizing a decision variable and marginalizing a chance variable from a mixed potential.

Suppose $\mu = (\alpha, \zeta, \upsilon)$ is a mixed potential for $r \cup s \cup t$, where $\alpha$ is a discrete potential for $r$, $\zeta$ is a continuous potential for $s$, and $\upsilon$ is a utility potential for $t$. Suppose $X \in \mathcal{D}$ and $X \in r \cup s \cup t$. Then the marginal of $\mu$ by deleting $X$, denoted by $\mu^{-X}$, is defined as follows:

$$\mu^{-X} = \begin{cases} (\alpha, \zeta, \upsilon^{-X}) & \text{if } X \notin r, X \notin s, \text{ and } X \in t \\ (\iota_d, \zeta, (\alpha \otimes \upsilon)^{-X}) & \text{if } X \in r, X \notin s, \text{ and } X \in t \\ (\alpha, \iota_c, (\zeta \otimes \upsilon)^{-X}) & \text{if } X \notin r, X \in s, \text{ and } X \in t \\ (\iota_d, \iota_c, (\alpha \otimes \zeta \otimes \upsilon)^{-X}) & \text{if } X \in r, X \in s, \text{ and } X \in t \end{cases} \quad (9)$$

We will assume that each decision variable is the domain of at least one utility potential. If so, at the time when a decision variable is to be marginalized from a mixed potential, it will always be in the domain of the utility part. Thus, we have only four cases in eq. (9).

Suppose $\mu = (\alpha, \zeta, \upsilon)$ is a mixed potential for $r \cup s \cup t$, where $\alpha$ is a discrete potential for $r$, $\zeta$ is a continuous potential for $s$, and $\upsilon$ is a utility potential for $t$. Suppose $X \in \mathcal{C}$ and $X \in r \cup s \cup t$. Then the marginal of $\mu$ by deleting $X$, denoted by $\mu^{-X}$, is defined as follows:

$$\mu^{-X} = ((\alpha \otimes \zeta)^{-X}, \iota_c, ((\alpha \otimes \zeta) \oslash (\alpha \otimes \zeta)^{-X}) \otimes \upsilon)^{-X})$$
$$\quad \text{if } X \in r, X \in s, X \in t, \text{ and } (r \cup s) \setminus \{X\} \subseteq \mathcal{C}_d,$$
$$= (\iota_d, (\alpha \otimes \zeta)^{-X}, ((\alpha \otimes \zeta) \oslash (\alpha \otimes \zeta)^{-X}) \otimes \upsilon)^{-X})$$
$$\quad \text{if } X \in r, X \in s, X \in t, \text{ and } (r \cup s) \setminus \{X\} \nsubseteq \mathcal{C}_d,$$
$$= (\alpha, \zeta^{-X}, ((\zeta \oslash \zeta^{-X}) \otimes \upsilon)^{-X})$$
$$\quad \text{if } X \notin r, X \in s, X \in t, \text{ and } s \setminus \{X\} \nsubseteq \mathcal{C}_d,$$
$$= (\alpha \otimes \zeta^{-X}, \iota_c, ((\zeta \oslash \zeta^{-X}) \otimes \upsilon)^{-X})$$
$$\quad \text{if } X \notin r, X \in s, X \in t, \text{ and } s \setminus \{X\} \subseteq \mathcal{C}_d,$$
$$= (\alpha^{-X}, \zeta, ((\alpha \oslash \alpha^{-X}) \otimes \upsilon)^{-X})$$
$$\quad \text{if } X \in r, X \notin s, X \in t, \text{ and } r \setminus \{X\} \nsubseteq \mathcal{C}_c,$$
$$= (\iota_d, \alpha^{-X} \otimes \zeta, ((\alpha \oslash \alpha^{-X}) \otimes \upsilon)^{-X})$$
$$\quad \text{if } X \in r, X \notin s, X \in t, \text{ and } r \setminus \{X\} \subseteq \mathcal{C}_c,$$
$$= (\alpha, \zeta, \upsilon^{-X}) \quad \text{if } X \notin r, X \notin s, \text{ and } X \in t,$$
$$= ((\alpha \otimes \zeta)^{-X}, \iota_c, \upsilon) \quad \text{if } X \in r, X \in s, X \notin t,$$
$$\quad \text{and } (r \cup s) \setminus \{X\} \subseteq \mathcal{C}_d,$$
$$= (\iota_d, (\alpha \otimes \zeta)^{-X}, \upsilon) \quad \text{if } X \in r, X \in s, X \notin t,$$
$$\quad \text{and } (r \cup s) \setminus \{X\} \nsubseteq \mathcal{C}_d,$$
$$= (\alpha, \zeta^{-X}, \upsilon) \quad \text{if } X \notin r, X \in s, X \notin t,$$
$$\quad \text{and } s \setminus \{X\} \nsubseteq \mathcal{C}_d,$$
$$= (\alpha \otimes \zeta^{-X}, \iota_c, \upsilon) \quad \text{if } X \notin r, X \in s, X \notin t,$$
$$\quad \text{and } s \setminus \{X\} \subseteq \mathcal{C}_d,$$
$$= (\alpha^{-X}, \zeta, \upsilon) \quad \text{if } X \in r, X \notin s, X \notin t,$$
$$\quad \text{and } r \setminus \{X\} \nsubseteq \mathcal{C}_c,$$
$$= (\iota_d, \alpha^{-X} \otimes \zeta, \upsilon) \quad \text{if } X \in r, X \notin s, X \notin t,$$
$$\quad \text{and } r \setminus \{X\} \subseteq \mathcal{C}_c. \quad (10)$$

Some comments about the marginalization of a chance variable from a mixed potential are as follows. In the first six cases, we have division of potentials that correspond to arc reversal in influence diagrams [15, 3]. This is necessary when we have an additive factorization of the joint utility function since multiplication and addition are not associative [17]. The last six cases in which the chance variable being marginalized doesn't belong to the domain of the utility potential is exactly as discussed in [20].

The divisions in the first six cases of eq. (10) can be avoided if either there is no additive factorization of the joint utility function, i.e., there is a single utility potential in the ID representation, or if the divisor is a vacuous potential (i.e., a potential whose values are all ones). In either of these two cases, the definition of marginalization of a chance variable in eq. (10) simplifies as follows (only the cases where $X \in t$ are shown as the other cases remain unchanged):

$$\mu^{-X} = \begin{cases} (\iota_d, \iota_c, (\alpha \otimes \zeta \otimes \upsilon)^{-X}) & \text{if } X \in r, X \in s, X \in t, \\ (\alpha, \iota_c, (\zeta \otimes \upsilon)^{-X}) & \text{if } X \notin r, X \in s, X \in t, \\ (\iota_d, \zeta, (\alpha \otimes \upsilon)^{-X}) & \text{if } X \in r, X \notin s, X \in t. \end{cases} \quad (11)$$

**Solving Hybrid Influence Diagrams.** We have all the definitions needed to solve hybrid influence diagrams with deterministic variables. The solution algorithm is basically the same as described in [17, 11]. We use the Shenoy-Shafer architecture [18] to propagate the potentials in a join tree. All variables need to be marginalized in a sequence that respects the information constraints in the sense that if $X$ precedes $Y$ in the information sequence, then $Y$ must be marginalized before $X$. Each time we marginalize a decision variable, we keep track of where the maximum is attained (as a function of the remaining variables in the potential being marginalized). This yields a decision function for the decision variable. The collection of all decision functions constitutes an optimal strategy for the influence diagram.

## 3 Mixture of Polynomials Approximations

In this section, we describe MOP functions. [19] describes MOP approximations of the PDFs of the univariate normal and chi-square distribution, and the conditional linear Gaussian distribution in two dimensions.

### 3.1 MOP Functions

A one-dimensional function $f : \mathbb{R} \to \mathbb{R}$ is said to be a *mixture of polynomials* (MOP) function if it is a piecewise function of the form:

$$f(x) = \begin{cases} a_{0i} + a_{1i}x + \cdots + a_{ni}x^n & \text{for } x \in A_i, i = 1, \ldots, k, \\ 0 & \text{otherwise.} \end{cases} \quad (12)$$

where $A_1, \ldots, A_k$ are disjoint intervals in $\mathbb{R}$ that do not depend on $x$, and $a_{0i}, \ldots, a_{ni}$ are constants for all $i$. We will say that $f$ is a $k$-piece (ignoring the 0 piece), and $n$-degree (assuming $a_{ni} \neq 0$ for some $i$) MOP function.

The main motivation for defining MOP functions is that such functions are easy to integrate in closed form, and that they are closed under multiplication and integration. They are also closed under differentiation and addition.

An $m$-dimensional function $f : \mathbb{R}^m \to \mathbb{R}$ is said to be a MOP function if:

$$f(x_1, \ldots, x_m) = f_1(x_1) \cdot f_2(x_2) \cdots f_m(x_m) \quad (13)$$

where each $f_i(x_i)$ is a one-dimensional MOP function as defined in eq. (12). If $f_i(x_i)$ is a $k_i$-piece, $n_i$-degree MOP function, then $f$ is a $(k_1 \cdots k_m)$-piece, $(n_1 + \ldots + n_m)$-degree MOP function. Therefore it is important to keep the number of pieces and degrees to a minimum. [19, 21] discuss the process of finding MOP approximations of univariate and bivariate conditional distributions. For space considerations, these are not discussed here.

## 4 Two Examples

In this section, we illustrate our algorithm for solving hybrid influence diagram with deterministic variables by solving two problems. The first one is called Entrepreneur's problem [7], and has continuous chance and deterministic variables, a continuous decision variable, and an unfactored utility function. The second problem is an American put option [2]. This problem has continuous chance variables, discrete decision variables with continuous chance predecessors, and an additive factorization of the joint utility function.

### 4.1 Entrepreneur's Problem

This problem is adapted from [7]. An entrepreneur has to decide on a price for his product. When the entrepreneur selects a price $P$, the quantity $Q_n$ that he will sell is determined from the demand curve $Q_n(P)$. This quantity $Q_n$ will have a cost of manufacturing $C_n(Q_n)$ given by the total cost curve. The entrepreneur's profit will then be the difference between his revenue $P \cdot Q_n$ and his cost $C_n$ or $\pi = P \cdot Q_n - C_n$. The entrepreneur needs to find a price $P$ that will maximize his profit.

This problem would be very simple if the demand curve and total cost curve were known with certainty, but this is seldom the case. We shall assume that the quantity $Q_n(P)$ determined from the demand curve is only a nominal value and that the actual quantity sold will be $Q_n + Z_1$, where $Z_1$ is a standard normal random variable. Furthermore, producing this quantity

$Q_a = Q_n + Z_1$ will cost $C_a = C_n + Z_2$, where $Z_2$ is another independent standard normal random variable. Note that the profit is now $\pi = P \cdot Q_a - C_a$.

For the demand curve, the functional form is $Q_n(p) = \frac{\ln \alpha - \ln p}{\beta}$, where $0 < p \leq \alpha$, with parameters $\alpha = 50$, $\beta = \frac{1}{80}$. For the total cost function we assume the form $C_n(q_a) = k_0 + k_1 q_a + k_2(1 - e^{-k_3 q_a})$, with parameters $k_0 = 700$, $k_1 = 4$, $k_2 = 400$, $k_3 = \frac{1}{50}$. We restrict the range of $P$ to $[1, 47]$ to ensure that $Q_a$ is nonnegative (with a very high probability of 0.999999). An ID representation of the problem is depicted in Figure 1.

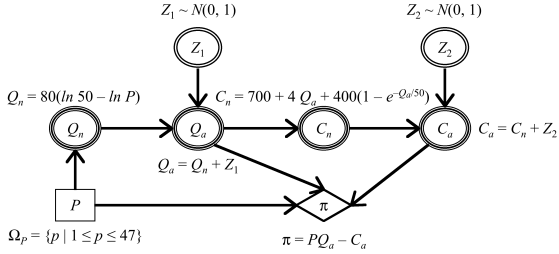

Figure 1: An ID representation of the entrepreneur's problem

We will solve the entrepreneur's problem by marginalizing variables in the following sequence: $C_a, Z_2, C_n, Q_a, Z_1, Q_n, P$. $Z_1$ and $Z_2$ are continuous variables with density potentials $\phi(z_1), \phi(z_2)$, where $\phi(\cdot)$ is the PDF of the standard normal random variable. As discussed in section 2, the conditional distributions of the deterministic variables $C_a, C_n, Q_a, Q_n$ are described by Dirac potentials. Since we have a single utility potential, no divisions are necessary during the solution process.

First, we marginalize $C_a$. The Dirac potential associated with $C_a$ is $\sigma_1(c_a, c_n, z_2) = \delta(c_a - (c_n + z_2))$. Resulting potential $\pi_1$ is a utility potential.

$$\begin{aligned}
\pi_1(p, q_a, c_n, z_2) &= (\pi \otimes \sigma_1)^{-C_a} \\
&= \int_{-\infty}^{\infty} \delta(c_a - (c_n + z_2))\,(pq_a - c_a)dc_a \\
&= p \cdot q_a - (c_n + z_2)
\end{aligned}$$

Next, we marginalize $Z_2$. Let $\varphi_{6,p}(z)$ denote the 6-piece, 3-degree MOP approximation of $\phi(z)$ as described in [21]. The density potential associated with $Z_2$ is $\varphi_{6,p}(z_2)$. The result of marginalizing $Z_2$ is the utility potential $\pi_2$ as follows.

$$\begin{aligned}
\pi_2(p, q_a, c_n) &= (\pi_1 \otimes \varphi_{6,p})^{-Z_2} \\
&= \int_{-\infty}^{\infty} \varphi_{6,p}(z_2)\,[p \cdot q_a - (c_n + z_2)]dz_2 \\
&= p \cdot q_a - c_n
\end{aligned}$$

Next we marginalize $C_n$. Let $f_{C_n}(q_a)$ denote the cost function $700 + 4q_a + 400(1 - e^{-\frac{q_a}{50}})$. Thus, the Dirac potential associated with $C_n$ is $\sigma_1(c_n, q_a) = \delta(c_n - f_{C_n}(q_a))$. There is no closed form for the result of marginalizing chance variable $C_n$. Therefore, we approximate $f_{C_n}(q_a)$ by a 3-piece, 3-degree, MOP approximation, that is denoted by $f_{pC_n}(q_a)$, in the interval $(2, 316)$ as follows:

$$f_{pC_n}(q_a) = \begin{cases} TSeries[f_{C_n}(q_a), q_a = 54, d = 3] & \text{if } 2 < q_a \leq 106 \\ TSeries[f_{C_n}(q_a), q_a = 158, d = 3] & \text{if } 106 < q_a \leq 201 \\ TSeries[f_{C_n}(q_a), q_a = 263, d = 3] & \text{if } 201 < q_a < 316 \\ 0 & \text{otherwise.} \end{cases} \quad (14)$$

The notation in eq. (14), introduced in [19], means using the Taylor series expansion of $f_{C_n}$ at a point $q_a$, up to degree $d$, in the specified interval. A graph of $f_{pC_n}$ overlaid on $f_{C_n}$ is shown in Figure 2.

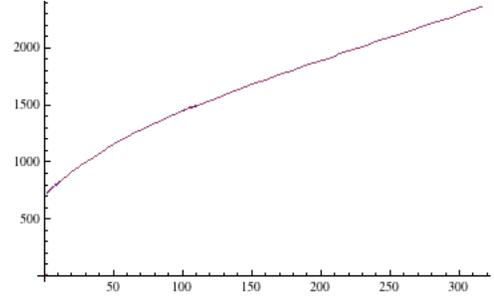

Figure 2: A graph of $f_{pC_n}$ overlaid on $f_{C_n}$

The result of marginalizing $C_n$ is denoted by $\pi_3$, which is a utility potential in MOP form.

$$\begin{aligned}
\pi_3(p, q_a) &= (\pi_2 \otimes \sigma_3)^{-C_n} \\
&= \int_{-\infty}^{\infty} \delta(c_n - f_{pC_n}(q_a))\,(pq_a - c_n)dc_n \\
&= p \cdot q_a - f_{pC_n}(q_a)
\end{aligned}$$

Next, we marginalize $Q_a$. The Dirac potential associated with $Q_a$ is $\sigma_4(q_a, q_n, z_1) = \delta(q_a - (q_n + z_1))$. The result of marginalizing $Q_a$ is denoted by $\pi_4$, which is a utility potential in MOP form.

$$\begin{aligned}
\pi_4(p, q_n, z_1) &= (\pi_3 \otimes \sigma_4)^{-Q_a} \\
&= \int_{-\infty}^{\infty} \pi_3(p, q_a)\,\delta(q_a - (q_n + z_1))dq_a \\
&= p \cdot (q_n + z_1) - f_{pC_n}(q_n + z_1)
\end{aligned}$$

Next, we marginalize $Z_1$. The density potential for $Z_1$ is $\varphi_{6,p}(z_1)$. The result of marginalizing $Z_1$ is denoted by $\pi_5$, which is a utility potential. Notice that since $\varphi_{6,p}$ and $f_{pC_n}$ are MOP functions, $\pi_5$ is also a MOP function (46 pieces, 3 degree).

$$\begin{aligned}\pi_5(p, q_n) &= (\pi_4 \otimes \varphi_{6,p})^{-Z_1} \\ &= \int_{-\infty}^{\infty} \pi_4(p, q_n, z_1)\, \varphi_{6,p}(z_1) dz_1\end{aligned}$$

Next, we marginalize $Q_n$. Let $f_{Q_n}$ denote the demand function: $f_{Q_n}(p) = \frac{\ln 50 - \ln p}{80}$, where $1 \leq p \leq 47$. We use a 3-piece, 3-degree MOP function $f_{pQ_n}$ to approximate $f_{Q_n}$.

$$f_{pQ_n}(p) = \begin{cases} TSeries[f_{Q_n}(p), p = 4, d = 3] & \text{if } 1 \leq p \leq 7 \\ TSeries[f_{Q_n}(p), p = 14, d = 3] & \text{if } 7 < p \leq 21 \\ TSeries[f_{Q_n}(p), p = 34, d = 3] & \text{if } 21 < p \leq 47 \\ 0 & \text{otherwise.} \end{cases}$$

The Dirac potential associated with $Q_n$ is $\sigma_5(q_n, p) = \delta(q_n - f_{pQ_n}(p))$. The result of marginalizing $Q_n$ is denoted by $\pi_6$, which is a utility potential for $P$ in MOP form.

$$\begin{aligned}\pi_6(p) &= (\pi_5 \otimes \sigma_5)^{-Q_n} \\ &= \pi_5(p, f_{pQ_n}(p))\end{aligned}$$

Figure 3 shows a graph of $\pi_6$ vs. $p$. Finally, we marginalize $P$. The maximum profit is $194.87 at $p = $24.40$. For comparison, when demand and supply are known with certainty, the problem reduces to a simple nonlinear optimization problem and the maximum profit $198 is obtained when price is $24.10. This completes the solution of the Entrepreneur's problem.

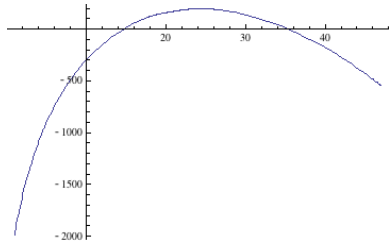

Figure 3: A graph of $\pi_6(p)$ vs. $p$

## 4.2 An American Put Option

This problem is adapted from [2]. An option trader has to decide whether to exercise or not a put option $p$ with initial stock price $S_0 = $40 and exercise price $X = $35. The option is available for exercise at three equally-spaced decision points over a 7-month period. Following standard practice in the financial literature, the stock prices, $S_1, S_2 \ldots S_k$ evolve according to the discrete stochastic process: $S_j = S_{j-1} \cdot Y$, where $Y \sim LN((r - \frac{\sigma^2}{2})\Delta t, \sigma^2 \Delta t)$, for $j = 1, 2, \ldots, k$, $S_j$ is the stock price at time $j\Delta t$, $r$ is the risk-less annual interest rate, $\sigma$ is the stock's volatility, $T$ denotes the length of the option (in years), and $\Delta t = \frac{T}{k}$. We assume $r = 0.0488$, $T = 0.5833$ years, $\Delta t = 0.1944$ years, $k = 3$ stages and $\sigma = 0.3$. Thus, $S_1 \sim LN(\ln 40 + 0.00074, 0.13229^2)$, $S_2|s_1 \sim LN(\ln s_1 + 0.00074, 0.13229^2)$, $S_3|s_2 \sim LN(\ln s_2 + 0.00074, 0.13229^2)$. An ID representation of the problem is shown in Figure 4.

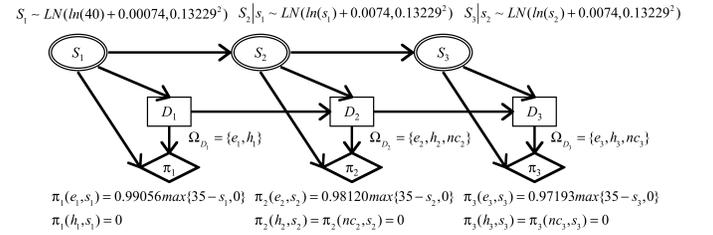

Figure 4: An ID representation of the American put option

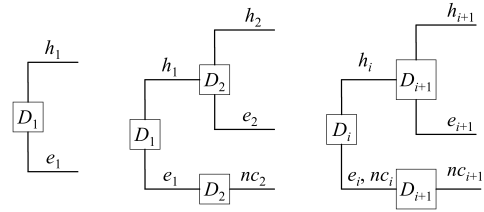

Figure 5: Conditionals for the put option decision nodes

The conditionals for the decision nodes in the problem are shown in Figure 5, where $e_i, h_i, nc_i$ denote the alternatives: exercise, hold, or no choice, respectively, for decision $D_i$. The only possible decision for stage $i$ is no choice if the stock was exercised at a prior time. The additive factors of the utility function are: $\pi_j = e^{-rj\Delta t} \max\{X - S_j, 0\}$, if $D_j = e_j$; $\pi_j = 0$, otherwise.

We approximate the marginal PDF of $S_1$ by a MOP function $\phi_1(s_1)$. Also the MoP approximations of the conditional PDFs for $S_2|s_1$, and $S_3|s_2$ are denoted by $\psi_1(s_1, s_2)$, and $\psi_2(s_2, s_3)$, respectively.

**Marginalizing $D_3$ and $S_3$.** Since no arc reversals are needed in this problem, no divisions are done, and we can use the definitions given in eq. (11). Since $\pi_3(e_3, s_3) \geq \pi_3(h_3, s_3)$ and $\pi_3(e_3, s_3) \geq \pi_3(nc_3, s_3)$,

the marginalized utility function is:

$$\pi'_3(d_2, s_3) = \begin{cases} \pi_3(e_3, s_3) & \text{if } d_2 = h_2, \\ 0 & \text{otherwise.} \end{cases}$$

Thus the optimal strategy for $D_3$ is to always exercise the option (assuming this alternative is available, and $s_3 < 35$).

Marginalizing $S_3$ involves combination and marginalization:

$$\pi''_3(d_2, s_2) = \begin{cases} \int_{-\infty}^{\infty} \pi'_3(d_2, s_3) \, \psi_2(s_2, s_3) ds_3 & \text{if } d_2 = h_2, \\ \pi_2(e_2, s_2) & \text{if } d_2 = e_2, \\ 0 & \text{otherwise.} \end{cases}$$

The ID after marginalizing $D_3$ and $S_3$ is shown in Figure 6.

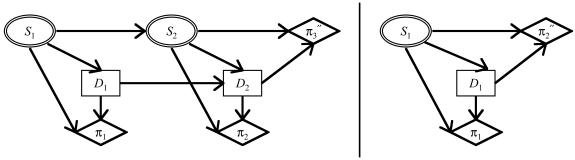

Figure 6: The ID after marginalizing $D_3$ and $S_3$ (left), and after marginalizing $D_2$ and $S_2$

**Marginalizing $D_2$ and $S_2$.** From Figure 7, we know that $\pi''_3(e_2, s_2) \geq \pi''_3(h_2, s_2)$, if $s_2 \leq 30.14$. Suppose $I(R)$ denotes the indicator function for the region $R$. Thus the marginalized utility function is:

$$\pi'_2(d_1, s_2) = \begin{cases} \pi''_3(e_2, s_2) \, I(0 \leq s_2 \leq 30.14) + \\ \pi''_3(h_2, s_2) \, I(s_2 \geq 30.14) & \text{if } d_1 = h_1, \\ 0 & \text{otherwise.} \end{cases}$$

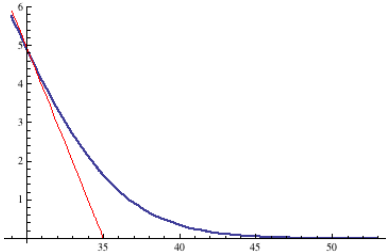

Figure 7: Utility functions $\pi''_3(h_2, s_2)$ (in bold) and $\pi''_3(e_2, s_2)$ vs. $s_2$

Optimal strategy for $D_2$ is to exercise the option if it is available and if the stock price is less than 30.14; otherwise hold it until $D_3$.

Marginalizing $S_2$ is similar to $S_3$:

$$\pi''_2(d_1, s_1) = \begin{cases} \int_{-\infty}^{\infty} \pi'_2(d_2, s_2) \, \psi_1(s_1, s_2) ds_2 & \text{if } d_1 = h_1, \\ \pi_1(e_1, s_1) & \text{if } d_1 = e_1, \\ 0 & \text{otherwise.} \end{cases}$$

**Marginalizing $D_1$ and $S_1$.** From Figure 8, we know that $\pi''_2(e_1, s_1) \geq \pi''_2(h_1, s_1)$, if $s_1 \leq 30.00$. Thus the marginalized utility function is:

$$\pi'_1(s_1) = \pi''_2(e_1, s_1) \, I(0 \leq s_1 \leq 30.00) + \\ \pi''_2(h_1, s_1) \, I(s_1 \geq 30.00)$$

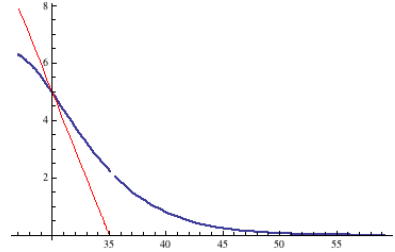

Figure 8: Utility functions $\pi''_2(h_1, s_1)$ (in bold) and $\pi''_2(e_1, s_1)$ vs. $s_1$

Optimal strategy for $D_1$ is to exercise the option when the stock price is less than 30.00; otherwise hold it until $D_2$.

The value of the option is $\pi''_1$:

$$\pi''_1(\diamond) = \int_{-\infty}^{\infty} \pi'_1(s_1) \, \phi_1(s_1) \, ds_1 = 1.19$$

Our result compares favorably to option pricing theory that prices it at \$1.219 and the result \$1.224 computed by Monte Carlo method using 30 stages [2]. This completes the solution of this problem.

## 5 Summary and Discussion

The main contribution of this paper is a framework for solving hybrid IDs with discrete, continuous, and deterministic chance variables, and discrete and continuous decision variables. The extended Shenoy-Shafer architecture for making inferences in hybrid Bayes nets has been extended to include decision variables and utility functions.

Two main problems in solving hybrid IDs are marginalization of continuous chance variables and marginalization of continuous decision variables. For decision problems that do not involve divisions, a solution is to approximate PDFs by MOP functions.

MOP functions are closed under multiplication, addition, integration, and differentiation. They are not closed under divisions. Thus MOP approximations could be used to mitigate the problems associated with marginalization of continuous and decision variables.